\documentclass[10pt,twocolumn,letterpaper]{article}

\usepackage{times}
\usepackage{epsfig}
\usepackage{graphicx}
\usepackage{amsmath}
\usepackage{amssymb}

\usepackage[pagebackref=true,breaklinks=true,letterpaper=true,colorlinks,bookmarks=false]{hyperref}

\usepackage[utf8]{inputenc} % allow utf-8 input
\usepackage[T1]{fontenc}    % use 8-bit T1 fonts
\usepackage{hyperref}       % hyperlinks
\usepackage{url}            % simple URL typesetting
\usepackage{booktabs}       % professional-quality tables
\usepackage{amsfonts}       % blackboard math symbols
\usepackage{nicefrac}       % compact symbols for 1/2, etc.
\usepackage{microtype}      % microtypography
\usepackage{times}
\usepackage{epsfig}
\usepackage{graphicx}
\usepackage{amsmath}
\usepackage{amssymb}
\usepackage{algorithmic}
\usepackage[linesnumbered,ruled,vlined]{algorithm2e}
\usepackage{multirow}
\usepackage{booktabs}
\usepackage{siunitx}
\usepackage{bm}
\usepackage{amsfonts}
\usepackage{color}

\begin{document}
	
	%%%%%%%%% TITLE
	\title{Discriminative Label Consistent  Domain Adaptation}
	
	\author{Lingkun Luo$^1$$^,$$^2$   Liming Chen $^2$  ,   Ying lu $^2$, Shiqiang Hu $^1$ \\
		Institution1:Shanghai Jiao Tong University\\
		Institution2: Laboratoire d'InfoRmatique en Image et Systèmes d'information / École Centrale de Lyon \\
		{\tt\small  lolinkun@gmail.com, liming.chen@ec-lyon.fr } }

	\maketitle
	%\thispagestyle{empty}
	
	%%%%%%%%% ABSTRACT
	\begin{abstract}

Domain adaptation (DA) is transfer learning which aims to learn an effective predictor on target data from source data despite  data distribution mismatch between source and target. We present in this paper a novel unsupervised DA method for cross-domain visual recognition which simultaneously optimizes the three terms of a theoretically established error bound. Specifically, the proposed DA method iteratively searches a latent shared feature subspace where not only the divergence of data distributions between the source domain and the target domain is decreased as most state-of-the-art DA methods do, but also the inter-class distances are increased to facilitate discriminative learning. Moreover, the proposed DA method sparsely regresses class labels from the features achieved in the shared  subspace while minimizing the prediction errors on the source data and ensuring label consistency between source and target. Data outliers are also accounted for to further avoid negative knowledge transfer.
Comprehensive experiments and in-depth analysis verify the effectiveness of the proposed DA method which consistently outperforms the state-of-the-art DA methods on standard DA benchmarks, \textit{i.e.},  12 cross-domain image classification tasks.

	\end{abstract}
	
\vspace{-5pt}
	\section{Introduction}
\vspace{-2pt}
	Traditional machine learning tasks assume that both training and testing data are drawn from a same data distribution\cite{pan2010survey,7078994}. However, in many real-life applications, due to different factors as diverse as sensor difference, lighting changes, viewpoint variations, \textit{etc.}, data from a target domain may have a different data distribution with respect to the labeled data in a source domain where a predictor can be reliably learned. On the other hand, manually labeling enough target data for the purpose of training an effective predictor can be very expensive, tedious and thus prohibitive.  
    
    Domain adaptation (DA) \cite{pan2010survey,7078994} aims to leverage possibly abundant labeled data from a \textit{source} domain to learn an effective predictor for data in a \textit{target} domain despite the data distribution discrepancy between the source and target. While DA can be \textit{semi-supervised} by assuming a certain amount of labeled data is available in the target domain, in this paper we are interested in \textit{unsupervised} DA where we assume that the target domain has no labels.

%     Based on this assumption, a series previous research designed to solve those problems, which have solid theoretical supports via , \textit{e.g.}, Hoeffding's Inequality\cite{cesa2004generalization}, Convex Optimization\cite{boyd2004convex}. We could calculate an adaptive classifier $h$ from hypothesis set ${{\mathcal H}}$, which supposes to reduce mismatches  in training data $min\left| {{E_{train}}(h)} \right|$ and error exist in test dataset $min\left| {{E_{train}}(h) - {E_{test}}(h)} \right|$ simultaneously. However, this assumption does not hold in many applications. For example, in face recognition, the distribution between training and test data can be discrepant due to facial gesture variation, illumination changes and facial occlusions. On the other hand, manual annotation of large training data could be extremely tedious\cite{cesa2004generalization,Gholami_2017_ICCV} and prohibitive for a given application. An interesting solution to this problem is to leverage abundant existing labeled data from a different but related domain (source domain) and generalize a predictive model learned from the source domain to unlabeled target data (target domain) despite the discrepancy between the source and target data distributions.
	
% 	Fortunately, domain adaptation(DA) techniques\cite{pan2010survey,7078994} try to solve the distribution mismatch problems and learn a domain-invariant predictive model from data. 
    
    Recent DA methods can be categorized into instance-based adaptation\cite{pan2010survey,donahue2013semi} and feature-based adaptation\cite{long2013transfer,DBLP:journals/tip/XuFWLZ16}. The instance-based approach generally assumes that   1) the conditional distributions of source and target domain are identical\cite{Zhang_2017_CVPR}, and 2) certain portion of the data in the source domain can be reused\cite{pan2010survey} for learning in the target domain through reweighting. Feature based adaptation relaxes such a strict assumption and only requires     
 that there exists a mapping from the  input data space to a latent shared feature representation space. This latent shared feature space captures the information necessary for training classifiers for source and target tasks. In this paper, we propose a feature-based adaptation DA method. 
 
A common method to approach feature adaptation is to seek a low-dimensional latent subspace\cite{7078994,Busto_2017_ICCV} via dimension reduction.  State of the art  features two main lines of approaches, namely data geometric structure alignment-based or data distribution centered. Data geometric structure alignment-based approaches, \textit{e.g.}, LTSL\cite{DBLP:journals/ijcv/ShaoKF14} , LRSR \cite{DBLP:journals/tip/XuFWLZ16},  seek a subspace where source and target data can be well aligned and interlaced in preserving inherent hidden geometric data structure via low rank constraint and/or sparse representation.  Data distribution centered methods aim to search a latent subspace where the discrepancy between the source and target data distributions is minimized, via various distances, \textit{e.g.}, Bregman divergence\cite{si2010bregman} based distance, Geodesic distance\cite{gong2012geodesic} or Maximum Mean Discrepancy\cite{gretton2007kernel} (MMD). The most popular distance is MMD due to its simplicity and solid theoretical foundations. 

% \ying{... State-of-the-art works feature two main lines of approaches, namely ... and ...}

A cornerstone theoretical result in DA \cite{ben2007analysis,ben2010theory,kifer2004detecting} is achieved by  Ben-David \textit{et al.}, who estimated an error bound of a learned hypothesis $h$ on a  target domain:  
\vspace{-5pt} 
\begin{equation}\label{eq:bound}
		\resizebox{0.90\hsize}{!}{%
			$\begin{array}{l}
			{e_{\cal T}}(h) \le {e_{\cal S}}(h) + {d_{\cal H}}({{\cal D}_{\cal S}},{{\cal D}_{\cal T}})+ \\
			\;\;\;\;\;\;\;\; \min \left\{ {{{\cal E}_{{{\cal D}_{\cal S}}}}\left[ {\left| {{f_{\cal S}}({\bf{x}}) - {f_{\cal T}}({\bf{x}})} \right|} \right],{{\cal E}_{{{\cal D}_{\cal T}}}}\left[ {\left| {{f_{\cal S}}({\bf{x}}) - {f_{\cal T}}({\bf{x}})} \right|} \right]} \right\}
			\end{array}$}
	\end{equation}
	\vspace{-10pt} 
	
	Eq.(\ref{eq:bound})  provides insight on the way to improve DA algorithms as it states that the performance of a hypothesis 
	$h$ on a target domain is determined by: 1) the classification error on the source domain ${e_{\cal S}}(h)$; 2) ${{d_{\cal H}}({{\cal D}_{\cal S}},{{\cal D}_{\cal T}})}$ which measures the $\mathcal{H}$\emph{-divergence}\cite{kifer2004detecting} between two distributions($\mathcal{D_S}$, $\mathcal{D_T}$); 3) the difference in labeling functions across the two domains. In light of this theoretical result, we can see that  data distribution centered DA methods only seek to minimize the second term in reducing data distribution discrepancies,  whereas data geometric structure alignment-based methods account for the underlying data geometric structure and expect but without theoretical guarantee the alignment of data distributions.

	Different from state of the art DA methods, we propose in this paper a novel Discriminative Label Consistent DA (DLC-DA) method which provides  a unified framework for a simultaneous optimization of the three terms in the  upper-bound error in Eq.(\ref{eq:bound}).  Specifically, the proposed DLC-DA also seeks a latent feature subspace to align data distributions as other state of the art methods , \textit{e.g.}, TCA\cite{pan2011domain}, JDA\cite{long2013transfer}, do, but also introduces a \textit{repulsive force} term in the proposed model so as to increase inter-class distances and thereby facilitate discriminative learning. More importantly, the proposed DLC-DA leverages existing labels in the source domain and ensures \textit{label consistencies} between the source and target domain through an iterative integrated linear label regression, thereby minimizing jointly the first and third term of the  error bound of the underlying learned hypothesis on the target domain.     
    
%     simultaneously  which extracts a latent shared feature space underlying the domains while reduces the distribution divergence between domain/sub-domain with same labels and increase divergence between domain/sub-domain with different labels simultaneously. Specifically, we learn a rescaled linear regression\cite{DBLP:conf/ijcai/0004YNH17} based constraint to better utilize prior knowledge. The extracted latent subspace and extracted feature of each samples can be optimized iteratively. Within each iteration, 1) the requirements of distribution divergence between domain/sub-domain with same or different labels could be satisfied, 2) discriminative attribute of extract subspace could be enhanced within well prediction proposed on target domain via rescaled linear regression and 3) the performance of label propagation increased through solving the unified framework.
	
    Comprehensive experiments carried out on standard DA benchmarks, \textit{i.e.},  12 cross-domain image classification tasks, verify the effectiveness of the proposed method, which consistently outperforms the state-of-the-art methods. In-depth analysis using both synthetic data and two 	additional partial models further provides insight of the proposed DA model and highlight its interesting properties. 
    
% 	Proposed method is different from sparse reconstruction based approaches\cite{DBLP:journals/ijcv/ShaoKF14,DBLP:journals/tip/XuFWLZ16} which require a strong assumption that target domain could be reconstructed by source domain once they project into a common subspace. This requirement would failed when source domain and target domain contain huge divergence. Moreover, those methods lack of theoretic guarantee that aligned source and target data have similar data distribution. Different from distance reduction based approaches\cite{DBLP:journals/pami/GhifaryBKZ17,long2013transfer}, we jointly consider reliable label propagation and divergence between domain/sub-domains, thereby avoiding negative knowledge transfer from the source domain. In addition, proposed Distribution Approximation and Rescaled Linear Regression domain adaptation (DRDA) is not a feature extraction technique but a DA technique. It can improve the performance with improved quality of features and easily combine fruits gained on deep learning research. Finally, we propose a simple yet effective algorithm with proven convergence to improve DA and feature selection simultaneously. A series of experiments have been performed on image classification tasks. The experimental results demonstrate competitive performance of proposed method.
	
The paper is organized as follows. Section 2 discusses the related work. Section 3 presents the method. Section 4 benchmarks the proposed DA method and provides in-depth analysis. Section 5 draws conclusion.   

	\section{Related work}

	Unsupervised Domain Adaptation (DA) assumes no labeled data are provided in the target domain. In earlier days this problem\cite{pan2010survey} is also known as co-variant shift and can be solved by  sample re-weighting. However, those methods fail when the divergence between the source and target domain becomes significant.
    
%     Unsupervised Domain Adaptation assumes no labeled data are provided in the target domain. Thus in order to achieve satisfactory classification performance on the target domain, one need to learn a classifier with labeled samples provided only from the source domain and unlabelled samples from the target domain.  In earlier days this problem\cite{pan2010survey} is also known as co-variant shift and can be solved by  sample re-weighting. However, those methods fail when the divergence between source and target domain becomes significant.
	
	Recent DA methods follow a mainstream approach which is based on \textit{feature adaptation}. The core idea of these methods is to search a latent shared feature space where both source and target data are statistically aligned, thereby a hypothesis learned using labeled source data can be an effective predictor for the unlabeled target data. Eq.(\ref{eq:bound}) by  Ben-David \textit{et al.} \cite{ben2007analysis,ben2010theory,kifer2004detecting} provides a theoretical foundation of this approach. In minimizing the divergence of data distributions between the source and target domain, feature adaptation-based DA methods decrease the second term of the upper error bound in Eq.(\ref{eq:bound}) and thereby improve performance of the learned predictor on the target domain. 
    
    The literature in feature adaptation has so far featured two main research lines: data distribution convergence (DDC)-based or data geometric structure alignment (DGSA)-based. In DDC-based DA methods, one aims to seek a latent shared feature subspace where the disparity of data distribution between source and target is minimized. For example,\cite{si2010bregman} proposed a Bregman Divergence based regularization schema, which combines Bregman divergence with conventional dimensionality reduction algorithms. In TCA\cite{pan2011domain}, the authors use a similar dimensionality reduction framework while making use of MMD to minimize the marginal distribution shift. JDA\cite{long2013transfer} goes one step further and proposes to simultaneously minimize  the discrepancies of the marginal and conditional distributions between source and target. Mahsa\cite{JMLR:v17:15-207} proposes a novel dimension reduction DA method via learning two different distances to compare the source and target distributions: the Maximum Mean Discrepancy and the Hellinger distance.  In DGSA-based DA methods, one seeks a shared feature subspace where target data can be sparsely reconstructed from source data \cite{DBLP:journals/ijcv/ShaoKF14} or source and target data are interleaved \cite{DBLP:journals/tip/XuFWLZ16}.
    
    In light of the three terms in the upper error bound defined by the right hand side of Eq.(\ref{eq:bound}), an optimized hypothesis on the target domain should simultaneously 1) minimize the prediction errors on the source domain, 2) decrease the divergence of data distributions and 3) ensure label consistency between the source and target. However, state of the art feature adaptation-based DA methods have only focused so far on data alignment, either statistically or geometrically. DDC-based DA methods only focus on bringing data distributions closer and may fall short to capture the inherent underlying data geometric structure. DGSA-based DA methods align data geometric structures between source and target in the searched feature subspace, and expect but without theoretical guarantee that the discrepancy of data distributions between source and target is implicitly reduced in the resultant subspace. As such, an interesting  move recently  is SCA \cite{DBLP:journals/pami/GhifaryBKZ17} and  JGSA \cite{Zhang_2017_CVPR} which jointly leverage data statistical and geometric properties in the search of the latent shared feature subspace. As shown in Fig.\ref{fig:compare} using synthetic data, another major disadvantage of most state of the art DA methods is that they do not consider discriminative knowledge hidden in the conditional distributions and as such they fall short to provide discriminativeness of data in the resultant feature subspace.

	In contrast to those previous DA methods, the proposed DLC-DA seeks a latent feature subspace which simultaneously optimizes the three terms of the upper error bound in Eq.(\ref{eq:bound}). Specifically, the proposed DLC-DA iteratively searches a shared feature subspace where 1) the source and target data distributions are discriminatively matched, and 2) the source and target labels can be sparsely regressed, thereby minimizing the prediction errors on the source data and ensuring label consistencies between the source and target domain. 

	\vspace{-2pt} 
	\section{The proposed method}
	\vspace{-2pt} 
% 	In this section, we present in detail the DLC-DA method for effective \textit{unsupervised}  domain adaptation.
	
	\subsection{Notations and Problem Statement}
	
	Matrices are written as boldface uppercase letters. Vectors are written as boldface lowercase letters. For matrix ${\bf{M}} = ({m_{ij}})$, its $i$-th row is denoted as ${{\bf{m}}^i}$, and its $j$-th column is denoted by ${{\bf{m}}_j}$.  We define the Frobenius norm ${\left\| . \right\|_F}$ and ${l_{2,1}}$ norm as: ${\left\| {\bf{M}} \right\|_F} = \sqrt {\sum {_{i = 1}^n} \sum {_{j = 1}^m} m_{ij}^2} $ and ${\left\| {\bf{M}} \right\|_{2,1}} = \sum {_{i = 1}^n} \sqrt {\sum {_{j = 1}^m} m_{ij}^2} $. 
    
    A domain $D$ is defined as an m-dimensional feature space $\chi$ and a marginal probability distribution $P(x)$, \textit{i.e.}, $\mathcal{D}=\{\chi,P(x)\}$ with $x\in \chi$.  Given a specific domain $D$, a  task $T$ is composed of a C-cardinality label set $\mathcal{Y}$  and a classifier $f(x)$,\textit{ i.e.}, $T = \{\mathcal{Y},f(x)\}$, where $f({x}) = \mathcal{Q}( y |x)$ can be interpreted as the class conditional probability distribution for each input sample $x$.

	In unsupervised domain adaptation, we are given a source domain $\mathcal{D_S}=\{x_{i}^{s},y_{i}^{s}\}_{i=1}^{n_s}$ with $n_s$ labeled samples ${{\bf{X}}_{\cal S}} = [x_1^s...x_{{n_s}}^s]$, which are associated with their class labels ${{\bf{Y}}_S} = {\{ {y_1},...,{y_{{n_s}}}\} ^T} \in {{\bf{\mathbb{R}}}^{{n_s} \times c}}$, and an unlabeled target domain $\mathcal{D_T}=\{x_{j}^{t}\}_{j=1}^{n_t}$ with $n_t$  unlabeled samples ${{\bf{X}}_{\cal T}} = [x_1^t...x_{{n_t}}^t]$, whose labels are ${{\bf{Y}}_T} = {\{ {y_{{n_s} + 1}},...,{y_{{n_s} + {n_t}}}\} ^T} \in {{\bf{\mathbb{R}}}^{{n_t} \times c}}$ are unknown. Here, ${y_i} \in {{\bf{\mathbb{R}}}^c}(1 \le i \le {n_s} + {n_t})$ is a binary vector in which $y_i^j = 1$ if ${x_i}$ belongs to the $j$-th class. We  define the data matrix ${\bf{X}} = [{{\bf{X}}_S},{{\bf{X}}_T}] \in {R^{m*n}}$ in packing both the source and target data. The source domain $\mathcal{D_S}$ and target domain $\mathcal{D_T}$ are assumed to be different, \textit{i.e.},  $\mathcal{\chi}_S=\mathcal{{\chi}_T}$, $\mathcal{Y_S}=\mathcal{Y_T}$, $\mathcal{P}(\mathcal{\chi_S}) \neq \mathcal{P}(\mathcal{\chi_T})$, $\mathcal{Q}(\mathcal{Y_S}|\mathcal{\chi_{S}}) \neq \mathcal{Q}(\mathcal{Y_T}|\mathcal{\chi_{T}})$.

	We also define the notion of \textit{sub-domain}, denoted as ${\cal D}_{\cal S}^{(c)}$, representing the set of samples in ${{\cal D}_{\cal S}}$ with the label $c$. Similarly, a sub-domain ${\cal D}_{\cal T}^{(c)}$ can be defined for the target domain as the set of samples in ${{\cal D}_{\cal T}}$ with the label $c$. However, as samples in the target domain ${{\cal D}_{\cal T}}$ are unlabeled, the definition of sub-domains in the target domain, requires a base classifier,\textit{ e.g.}, Nearest Neighbor (NN),  to attribute  pseudo labels for samples in ${{\cal D}_{\cal T}}$.

	The  maximum mean discrepancy (MMD)  is an effective non-parametric distance-measure  that compares the distributions of two sets of data by mapping the data to Reproducing Kernel Hilbert Space\cite{borgwardt2006integrating} (RKHS). Given two distributions $\mathcal{P}$ and $\mathcal{Q}$, the MMD between $\mathcal{P}$ and $\mathcal{Q}$ is defined as:
	\begin{equation}
		\label{eq:MMD}
		Dist(P,Q) = \parallel \frac{1}{n_1} \sum^{n_1}_{i=1} \phi(p_i) - \frac{1}{n_2} \sum^{n_2}_{i=1} \phi(q_i) \parallel_{\mathcal{H}}
	\end{equation}
	where $P=\{ p_1, \ldots, p_{n_1} \}$ and $Q = \{ q_1, \ldots, q_{n_2} \}$ are two random variable sets from distributions $\mathcal{P}$ and $\mathcal{Q}$, respectively, and $\mathcal{H}$ is a universal RKHS with the reproducing kernel mapping $\phi$: $f(x) = \langle \phi(x), f \rangle$, $\phi: \mathcal{X} \to \mathcal{H}$.

	The aim of the DLC-DA is to search jointly a transformation matrix ${\bf{A}} \in {R^{m*k}}$ projecting discriminatively both the source and target data into a  latent shared feature subspace and a label regressor in  simultaneously minimizing  the three terms of the upper error bound in Eq.(\ref{eq:bound}). 
	
	\subsection{Formulation}
	Specifically, we aim to define an integrated optimization model  with the following properties: P1) The classification error on the source domain is minimized; P2) the discrepancy between the two distributions ($\mathcal{D_S}$, $\mathcal{D_T}$) is reduced via ${{d_{\cal MMD}}({{\cal D}_{\cal S}},{{\cal D}_{\cal T}})}$ ; P3) inter-class distances in both the two domains  are increased so as to facilitate discriminative learning; P4):  label consistency across the two domains is explicitly maximized  through  iterative linear label regression,  and P5) Data outliers are accounted for to avoid negative transfer.
	\vspace{-2pt} 
	\subsubsection{Matching Marginal and Conditional Distributions}
	\vspace{-2pt} 
% 	Ben-David pointed out\cite{ben2007analysis,ben2010theory}, that the error of ${e_{\cal T}}(h)$ could be bound by ${d_{\cal H}}({{\cal D}_{\cal S}},{{\cal D}_{\cal T}})$\cite{kifer2004detecting} which measures the distribution divergence between ${{\cal D}_{\cal S}}$ and ${{\cal D}_{\cal T}}$ . 
    To meet property P2, we follow JDA\cite{long2013transfer} and explicitly leverage MMD in RKHS to measure the distances between the expectations of the source domain/sub-domain and target domain/sub-domain: \textbf{1)} The empirical distance of the source and target domains are defined as $Dis{t^{m}}$. \textbf{2)} The conditional distance $Dis{t^{c}}$ is defined as the sum of the empirical distances between sub-domains in ${{\cal D}_{\cal S}}$ and ${{\cal D}_{\cal T}}$ with a same label $c$.
    \vspace{-7pt}    \begin{equation}\label{eq:JDA}
		\resizebox{0.85\hsize}{!}{%
			$\begin{array}{l}
			Dis{t_C} = Dis{t^m}({D_S},{D_T}) + Dis{t^c}\sum\limits_{c = 1}^C {({D_S}^c,{D_T}^c)} \\
			\;\;\;\;\;\;\;\;\; = {\left\| {\frac{1}{{{n_s}}}\sum\limits_{i = 1}^{{n_s}} {{{\bf{A}}^T}{x_i} - } \frac{1}{{{n_t}}}\sum\limits_{j = {n_s} + 1}^{{n_s} + {n_t}} {{{\bf{A}}^T}{x_j}} } \right\|^2}\\
			\;\;\;\;\;\;\;\;\; + {\left\| {\frac{1}{{n_s^{(c)}}}\sum\limits_{{x_i} \in {D_S}^{(c)}} {{{\bf{A}}^T}{x_i}}  - \frac{1}{{n_t^{(c)}}}\sum\limits_{{x_j} \in {D_T}^{(c)}} {{{\bf{A}}^T}{x_j}} } \right\|^2}\\
			\;\;\;\;\;\;\;\;\; = tr({{\bf{A}}^T}{\bf{X}}({{\bf{M}}_{\bf{0}}} + \sum\limits_{c = 1}^{c = C} {{{\bf{M}}_c}} ){{\bf{X}}^{\bf{T}}}{\bf{A}})
			\end{array}$}
	\end{equation}
	
	where $C$ is the number of classes, $\mathcal{D_S}^{(c)} = \{ {x_i}:{x_i} \in \mathcal{D_S} \wedge y({x_i} = c)\} $ represents the ${c^{th}}$ sub-domain in the source domain. $\mathcal{D_T}^{(c)}$ is defined similarly for the target domain. ${{\bf{M}}_0}$ represents the marginal distribution between ${{\cal D}_{\cal S}}$ and ${{\cal D}_{\cal T}}$with ${{{({{\bf{M}}_0})}_{ij}} = \frac{1}{{{n_s}{n_s}}}}$ if $({{x_i},{x_j} \in {D_S}})$, ${{{({{\bf{M}}_0})}_{ij}} = \frac{1}{{{n_t}{n_t}}}}$ if $({{x_i},{x_j} \in {D_T}})$ and ${{{({{\bf{M}}_0})}_{ij}} = 0}$ otherwise. $\bf M_c$ represents the conditional distribution between the sub-domains $c$ in ${{\cal D}_{\cal S}}$ and ${{\cal D}_{\cal T}}$ with ${{{({{\bf{M}}_c})}_{ij}} = \frac{1}{{n_s^{(c)}n_s^{(c)}}}}$ if $({{x_i},{x_j} \in {D_S}^{(c)}})$, ${{{({{\bf{M}}_c})}_{ij}} = \frac{1}{{n_t^{(c)}n_t^{(c)}}}}$ if $({{x_i},{x_j} \in {D_T}^{(c)}})$, ${{{({{\bf{M}}_c})}_{ij}} = \frac{{ - 1}}{{n_s^{(c)}n_t^{(c)}}}}$ if $({{x_i} \in {D_S}^{(c)},{x_j} \in {D_T}^{(c)}\;or\;\;{x_i} \in {D_T}^{(c)},{x_j} \in {D_S}^{(c)}})$ and ${{{({{\bf{M}}_c})}_{ij}} = 0}$ otherwise.  $n_s^{(c)}$ is the number of samples in the ${c^{th}}$ {source} sub-domain and $n_t^{(c)}$ is defined similarly for  the ${c^{th}}$ {target} sub-domain .  
    
    The discrepancy between the marginal distributions $\mathcal{P}(\mathcal{X_S})$ and $\mathcal{P}(\mathcal{X_T})$ can be reduced in minimizing {$Dis{t^{m}}$} whereas the mismatches of conditional distributions between ${{D_{\cal S}}^c}$ and ${{D_{\cal T}}^c}$ can be decreased in minimizing ${Dis{t^{c}}}$. In summary,  the discrepancies of both the marginal and conditional distributions between the source and target  can be jointly reduced in minimizing $Dis{t_C}$. 
    
%     and ${d_{\cal MMD}}({{\cal D}_{\cal S}},{{\cal D}_{\cal T}})$ would minimized via decreasing of $Dis{t_C}$.
	
	\vspace{-5pt} 
	\subsubsection{Repulsing interclass data for discriminative DA}
	
	Aligning data distributions as does the previous sub-section does not guarantee that both the source and target data are discriminative with respect to the class labels.  To satisfy property P1 and P3,   we introduce a \textit{repulsive force} term $Dist_{{\cal S} \to {\cal S}}^{re}$,  where ${{\cal S} \to {\cal S}}$ indexes the distances computed from ${D_{\cal S}}$ to ${D_{\cal S}}$. $Dist_{{\cal S} \to {\cal S}}^{re}$ represents the sum of the distances from each source sub-domain ${D_{\cal S}}^{(c)}$ to all the other source sub-domains ${D_{\cal S}}^{(r);\;r \in \{ \{ 1...C\}  - \{ c\} \} }$, excluding the $c$-th source sub-domain:
%     	Aligning data distributions as does the previous sub-section does not guarantee that both the source and target data are discriminative with respect to the class labels.  To satisfy property P1 and P3,  Bound on target error ${e_{\cal T}}(h)$ could reduce via minimization of source error ${e_{\cal S}}(h)$. Inspired by LDA\cite{martinez2001pca}, performance of classifier $h$ on  ${{\cal D}_{\cal S}}$ could improve via extract discriminative information between different source sub-domains ${{D_{\cal S}}^c}$. To achieve this, we propose a discriminative constraint on  ${{\cal D}_{\cal S}}$ by  introduce \textit{repulsive force} $Dist_{{\cal S} \to {\cal S}}^{re}$, where ${{\cal S} \to {\cal S}}$ index the distances computed from ${D_{\cal S}}$ to ${D_{\cal S}}$. $Dist_{{\cal S} \to {\cal S}}^{re}$ represents the sum of the distances from each source sub-domain ${D_{\cal S}}^{(c)}$ to all the the source sub-domains ${D_{\cal S}}^{(r);\;r \in \{ \{ 1...C\}  - \{ c\} \} }$ except the source sub-domain with the label $c$, which defined as:3	\begin{equation}\label{eq:stos}
	\begin{equation}\label{eq:stos}
		\resizebox{0.85\hsize}{!}{%
			$\begin{array}{l}
			Dist_{S \to S}^{re} = Dis{t^c}\sum\limits_{c = 1}^C {({D_S}^c,{D_S}^{r \in \{ \{ 1...C\}  - \{ c\} \} })} \\
			\;\;\;\;\;\;\;\;\;\;\;\;\;\;\; = \sum\limits_{c = 1}^C {tr({{\bf{A}}^T}{\bf{X}}({{\bf{M}}_{S \to S}}){{\bf{X}}^{\bf{T}}}{\bf{A}})} 
			\end{array}$}
	\end{equation}
	
	Eq.(\ref{eq:stos}) is calculated in a similar way as Eq.(\ref{eq:JDA}), with  ${{{({{\bf{M}}_{{\bf{S}} \to {\bf{S}}}})}_{ij}} = \frac{1}{{n_s^{(c)}n_s^{(c)}}}}$ if ${({x_i},{x_j} \in {D_S}^{(c)})}$, ${ \frac{1}{{n_s^{(r)}n_s^{(r)}}}}$ if ${({x_i},{x_j} \in {D_S}^{(r)})}$, ${  \frac{{ - 1}}{{n_s^{(c)}n_s^{(r)}}}}$ if ${({x_i} \in {D_S}^{(c)},{x_j} \in {D_S}^{(r)}\;or\;{x_i} \in {D_S}^{(r)},{x_j} \in {D_S}^{(c)})}$ and ${ 0}$ otherwise. 

    Similarly, we can also introduce a \textit{repulsive force} term $Dist_{{\cal S} \to {\cal T}}^{re} + Dist_{{\cal T} \to {\cal S}}^{re}$ between the source sub-domains and those in the target domain, where ${{\cal S} \to {\cal T}}$ and ${{\cal T} \to {\cal S}}$ index the distances computed from ${D_{\cal S}}$ to ${D_{\cal T}}$ and those from ${D_{\cal T}}$ to ${D_{\cal S}}$, respectively.   $Dist_{{\cal S} \to {\cal T}}^{re}$ represents the sum of the distances between each source sub-domain ${D_{\cal S}}^{(c)}$ and all the  target sub-domains ${D_{\cal T}}^{(r);\;r \in \{ \{ 1...C\}  - \{ c\} \} }$ excluding the $c$-th target sub-domain.  $Dist_{{\cal T} \to {\cal S}}^{re}$ represents the sum of the distances from each target sub-domain ${D_{\cal T}}^{(c)}$ to all the the source sub-domains ${D_{\cal S}}^{(r);\;r \in \{ \{ 1...C\}  - \{ c\} \} }$ excluding the $c$-th  source sub-domain. These two distances are explicitly defined as:
	\begin{equation}\label{eq:CDDAnew}
		\resizebox{0.9\hsize}{!}{%	
			$\begin{array}{l}
			Dist_{S \to T}^{re} + Dist_{T \to S}^{re} = Dis{t^c}\sum\limits_{c = 1}^C {({D_S}^c,{D_T}^{r \in \{ \{ 1...C\}  - \{ c\} \} })} \\
			\;\;\;\;\;\;\;\;\;\;\;\;\;\;\;\;\;\;\;\;\;\;\;\;\;\;\;\;\;\;\;\;\; + Dis{t^c}\sum\limits_{c = 1}^C {({D_T}^c,{D_S}^{r \in \{ \{ 1...C\}  - \{ c\} \} })} \\
			\;\;\;\;\;\;\;\;\;\;\;\;\;\;\;\;\;\;\;\;\;\;\;\;\;\;\;\;\;\;\;\;\; = \sum\limits_{c = 1}^C {tr({{\bf{A}}^T}{\bf{X}}({{\bf{M}}_{S \to T}} + {{\bf{M}}_{T \to S}}){{\bf{X}}^{\bf{T}}}{\bf{A}})} 
			\end{array}$}
	\end{equation}
	
	Eq.(\ref{eq:CDDAnew}) is defined in a similar way as Eq.(\ref{eq:stos}) and Eq.(\ref{eq:JDA}), where ${{{({{\bf{M}}_{{\bf{S}} \to {\bf{T}}}})}_{ij}} = \frac{1}{{n_s^{(c)}n_s^{(c)}}}}$ if $({{x_i},{x_j} \in {D_S}^{(c)}})$, ${  \frac{1}{{n_t^{(r)}n_t^{(r)}}}}$ if $({{x_i},{x_j} \in {D_T}^{(r)}})$, ${  \frac{{ - 1}}{{n_s^{(c)}n_t^{(r)}}}}$ if $({{x_i} \in {D_S}^{(c)},{x_j} \in {D_T}^{(r)}\;or\;{x_i} \in {D_T}^{(r)},{x_j} \in {D_S}^{(c)}})$ and ${  0}$ otherwise; ${{{({{\bf{M}}_{{\bf{T}} \to {\bf{S}}}})}_{ij}} = \frac{1}{{n_t^{(c)}n_t^{(c)}}}}$ if ${({x_i},{x_j} \in {D_T}^{(c)})}$, ${  \frac{1}{{n_s^{(r)}n_s^{(r)}}}}$ if ${({x_i},{x_j} \in {D_S}^{(r)})}$, ${  \frac{{ - 1}}{{n_t^{(c)}n_s^{(r)}}}}$ if ${({x_i} \in {D_T}^{(c)},{x_j} \in {D_S}^{(r)}\;or\;{x_i} \in {D_S}^{(r)},{x_j} \in {D_T}^{(c)})}$ and ${  0}$ otherwise.
	
% 	Maximize the distances ($Dist_{{\cal S} \to {\cal T}}^{re} + Dist_{{\cal T} \to {\cal S}}^{re}$) could explicitly reduce conditional distribution $Q({{\cal D}_{\cal S}}^c|{{\cal Y}_{\cal T}} \ne c)\;and\;Q({{\cal D}_{\cal T}}^c|{{\cal Y}_{\cal S}} \ne c)$, which supposes to minimize difference in labeling function  across the two domains and bound more tightly the ${e_{\cal T}}(h)$.
	Finally, in integrating Eq.(\ref{eq:stos})  and Eq.(\ref{eq:CDDAnew}), we obtain the final repulsive force term as 
    	\vspace{-5pt} 
	\begin{equation}\label{eq:re}
		\resizebox{0.9\hsize}{!}{%
			${Dist}^{re} = \sum\limits_{c = 1}^C {tr({{\bf{A}}^T}{\bf{X}}({{\bf{M}}_{S \to T}} + {{\bf{M}}_{T \to S}} + {{\bf{M}}_{S \to S}}){{\bf{X}}^{\bf{T}}}{\bf{A}})} $}
	\end{equation}
	We further define ${{\bf{M}}_{REP}} = {{\bf{M}}_{S \to T}} + {{\bf{M}}_{T \to S}} + {{\bf{M}}_{S \to S}} $ as the \textit{repulsive force} constraint matrix. 
    
    The maximization of Eq.(\ref{eq:re})  increases the distances between source sub-domains as well as those between source and target sub-domains, thereby enhancing the discriminative power of the underlying latent feature space. 
		\vspace{-6pt} 	
	\subsubsection{Label Consistent Regression}
	
	  Eq.(\ref{eq:JDA})  and Eq.(\ref{eq:re}) do not explicitly optimize the prediction errors of a learned hypothesis on the source data nor ensure label consistencies between the source and target domain.    
     
%      We further optimize the latent subspace which extracted via MMD distribution measurement calculation through introducing a novel discriminant least square regression constraint $\Phi ({\bf{A}},{{\bf{Y}}_S},{{\bf{X}}_S})$, where \textbf{A} is the projection transformation.  We introduce this constraint due to its effectiveness for data analysis as well as its completeness in statistics theory.
     
     To  meet P1 and P4 and thereby explicitly optimize the first and the third term in Eq.(\ref{eq:bound}), we  introduce  a novel \textit{label regression consistency} constraint $\Phi ({\bf{A}},{{\bf{Y}}_S},{{\bf{X}}_S})$, where \textbf{A} is the transformation matrix projecting both the source and target data into a latent shared feature subspace of dimension $k$. Specifically, we first embed each $C$-dimensional label vector into the $k$-dimensional latent shared feature subspace by adding $(C-k+1)$ times $0$.   We can then perform class label  regression and explicitly enhance class prediction accuracy on the source data and  the class label  consistency  between the source and target domains through the least  square regression  (LSR): $\min \left\| {{{\bf{X}}^T}{\bf{A}} - {\bf{Y}}} \right\|_F^2$ ${\rm{s}}t.{\bf{Y}} \ge {\bf{0}},\;{\bf{Y1}} = {\bf{1}}$, with  \textbf{Y} the class label matrix as  defined in Section 3.1 and extended into a $n \times k$ matrix by embedding each label vector into a $k$-dimensional unit vector. This constraint simply expresses that each data sample should be projected in the vicinity of its corresponding unit label vector in the latent shared feature subspace. 
     
%      which suppose to transform data of both domains into a discriminant subspace in which the labeled knowledge can be well preserved and satisfy P1. \textbf{Y} is class label matrix, already defined in Section3.1. 
    
%     	MMD approaches support that data from different domains (${{\bf{X}}_S}$ and ${{\bf{X}}_T}$) could project into a unified distribution subspace $\text{A}$, via measuring the conditional and marginal distributions. Based on this assumption the least square regression (LSR) can be formulated as: $\min \left\| {{{\bf{X}}^T}{\bf{A}} - {\bf{Y}}} \right\|_F^2$ ${\rm{s}}t.{\bf{Y}} \ge {\bf{0}},\;{\bf{Y1}} = {\bf{1}}$, which suppose to transform data of both domains into a discriminant subspace in which the labeled knowledge can be well preserved and satisfy P1. \textbf{Y} is class label matrix, already defined in Section3.1. 
	
	As source and target data can be noisy, we also  introduce a matrix $\textbf{E}$ to model noise. As a result, the LSR can be reformulated as: $\min \left\| {{{\bf{X}}^T}{\bf{A}} - {\bf{Y}} + {\bf{E}}} \right\|_F^2$  ${\rm{s}}t.\;{\bf{Y}} \ge {\bf{0}},\;{\bf{Y1}} = {\bf{1}}$. The introduction of the error matrix $\textbf{E}$ enables to account for outliers and thereby alleviate the influence of negative transfer to meet property P4. 
	
	In many real-life applications, especially in the field of visual recognition, data of a given class generally lie within a manifold of much lower dimension in comparison with the original  data space, \textit{e.g.}, pixel number of images. Therefore, we further introduce a ${l_{2,1}}$-norm constraint so as to express the property that the class label of a data sample  should be regressed from a sparse combination of features in the latent shared feature subspace. This constraint introduces a regularization term on  $\textbf{A}$ for discriminative subspace projection. 
%     , since which is robust to outliers in data point and designed a joint sparsity subspace. Moreover,  ${l_{2,1}}$-norm defined in this model is convex and can be easily optimized. Then the problem can be further formulated as: $\min \left\| {{{\bf{X}}^T}{\bf{A}} - {\bf{Y}} - {\bf{E}}} \right\|_F^2 + \beta {\left\| {\bf{A}} \right\|_{2,1}}$ ${\rm{s}}t.{\bf{A}},e \succ 0,\;{\bf{Y}} \ge {\bf{0}},\;{\bf{Y1}} = {\bf{1}}$. 
	
% 	In this paper we propose $l_{2,1}^2$-norm constraint rather than ${l_{2,1}}$-norm. Since $\min \left\| {{{\bf{X}}^T}{\bf{A}} - {\bf{Y}} - 1{{\bf{e}}^T}} \right\|_F^2 + \beta \left\| {\bf{A}} \right\|_{2,1}^2$ equals with $\min \left\| {{{\bf{X}}^T}\Theta {{\bf{A}}^*} - {\bf{Y}} - 1{{\bf{e}}^T}} \right\|_F^2 + \beta \left\| {{{\bf{A}}^*}} \right\|_{2,1}^2$. ${\theta _j} = \left\| {{{\bf{A}}^j}} \right\|_2^2 \div \sum {_{j = 1}^d} \left\| {{{\bf{A}}^{j'}}} \right\|_2^2$ and ${{\bf{A}}^*} = \Theta {\bf{A}}$. ${\bf{e}}$ and $\theta $ represents row vectors of $\textbf{E}$ and $\Theta $. 
    The \textit{label regression consistency} constraint is finally  formulated as
	\begin{equation}\label{eq:RLR}
		\resizebox{0.6\hsize}{!}{%
			$\begin{array}{l}
			\min \left\| {{{\bf{X}}^T}{\bf{A}} - {\bf{Y}} + 1{{\bf{e}}^T}} \right\|_F^2 + \beta \left\| {\bf{A}} \right\|_{2,1}^2\\
			\;\;\;\;\;{\rm{s}}t.\;{\bf{Y}} \ge {\bf{0}},\;{\bf{Y1}} = {\bf{1}}
			\end{array}$}
	\end{equation}
	
% 	which could measure the contributes of features and inherit merits hidden behind the rescaled linear regression.
	
	\subsubsection{The final model}
	In integrating all the properties expressed in the previous subsections, \textit{i.e.}, Eq.(\ref{eq:JDA}), Eq.(\ref{eq:re}) and Eq.(\ref{eq:RLR}), we obtain our  final DA model, formulated as Eq.(\ref{eq:opti})
\begin{equation}\label{eq:opti}
		\resizebox{1\hsize}{!}{%
			$\begin{array}{*{20}{l}}
			{\mathop {\min }\limits_{{\bf{A}},{\bf{e}},{{\bf{Y}}_{\bf{U}}} \ge 0,{{\bf{Y}}_{\bf{U}}}{\bf{1}} = {\bf{1}}} (tr({{\bf{A}}^T}{\bf{X}}{{\bf{M}}^*}{{\bf{X}}^T}{\bf{A}}) + \alpha \left\| {\bf{A}} \right\|_F^2 + \beta \left\| {\bf{A}} \right\|_{2,1}^2)}\\
			{\;\;\;\;\;\;\;\;\;\;\;\;\;\;\;\;\;\;\;\;\;\;\;\;\;\; + \left\| {{{\bf{X}}^T}{\bf{A}} - {\bf{1}}{{\bf{e}}^T} - {\bf{Y}}} \right\|_F^2}\\
			{{\rm{s}}t.{{\bf{M}}^*} = {{\bf{M}}_0} + \sum _{c = 1}^C({{\bf{M}}_c}) - {{\bf{M}}_{REP}},\;{\bf{Y}} \ge {\bf{0}},\;{\bf{Y1}} = {\bf{1}}}
			\end{array}$}
	\end{equation}

	Through iterative optimization of Eq.(\ref{eq:opti}), our DA method searches jointly a latent subspace and a label regression model satisfying at the same time properties P1 through P5.
	
	\subsection{Solving the model}
	Eq.(\ref{eq:opti}) is not convex. We propose an effective method that solves each variable in a coordinate descent manner. Main steps for solving Eq.(\ref{eq:opti}) are as follows. All the key steps have a closed form solution:
	
	\textit{\textbf{Step.1}} (Initialization of ${{\bf{M}}^*}$) ${{\bf{M}}^*}$ can be initialized by calculating ${{\bf{M}}_0}$ since there is no labels or pseudo labels on the target domain yet. We obtain ${{\bf{M}}^*} = {{\bf{M}}_0}$ where ${{\bf{M}}_0}$ is defined as in Eq.(\ref{eq:JDA}).

	\textit{\textbf{Step.2}} (Initialization of ${\bf{A}}$) ${\bf{A}}$ can be initialized to reduce the marginal distributions between $\mathcal{P}(\mathcal{X_S})$ and $\mathcal{P}(\mathcal{X_T})$ and to calculate an adaptive subspace  via the Rayleigh quotient algorithm in solving Eq.(\ref{eq:S2}): 
\begin{equation}\label{eq:S2}
		\resizebox{0.6\hsize}{!}{%
			$({\bf{X}}{{\bf{M}}_0}{{\bf{X}}^T} + \alpha {\bf{I}}){\bf{A}} = {\bf{XH}}{{\bf{X}}^T}{\bf{A}}\Phi$}
	\end{equation}
	where ${\bf{H}}{\rm{ = }}{\bf{I}} - \frac{1}{n}{\bf{1}}$ is the centering matrix, ${\bf{X}} = {{\bf{X}}_S} \cup {{\bf{X}}_T}$ and $\Phi  = diag({\phi _1},...,{\phi _k}) \in {R^{k \times k}}$ are Lagrange multipliers.  ${\bf{A}}$ is then initialized as the $k$ smallest eigenvectors of Eq.(\ref{eq:S2}).

	\textit{\textbf{Step.3}} (Update of ${\bf{e}}$)  ${\bf{e}}$ is updated in solving Eq.(\ref{eq:opti}) with other variables held fixed. To update ${\bf{e}}$, one should solve  Eq.(\ref{eq:S41})	
 	\vspace{-5pt} 
 \begin{equation}\label{eq:S41}
		\resizebox{0.6\hsize}{!}{%
			${\bf{e'}} = \arg \mathop {\min }\limits_{\bf{e}} \left\| {{{\bf{X}}^T}{\bf{A}} + {\bf{1}}{{\bf{e}}^T} - {\bf{Y}}} \right\|_F^2$}
	\end{equation}

	In setting to $0$ the partial derivative of Eq.(\ref{eq:S41}) with respect to $\textbf{e}$, we achieve the optimal solution of $\textbf{e}$ as	
	\vspace{-5pt} 
\begin{equation}\label{eq:S42}
		\resizebox{0.43\hsize}{!}{%
			${\bf{e}} = \frac{1}{n}({{\bf{Y}}^T}{\bf{1}} - {{\bf{A}}^T}{\bf{X1}})$}
	\end{equation}

	\textit{\textbf{Step.4}} (Update of ${\bf{A}}$) ${\bf{A}}$ is updated by solving the optimization problem in Eq.(\ref{eq:opti}) with other variables held fixed. To make sure Eq.(\ref{eq:opti}) is differentiable, we regularize $\left\| {\bf{A}} \right\|_{2,1}^2$ as $(\sum {_{j = 1}^d} \sqrt {\left\| {{{\bf{a}}^j}} \right\|_2^2 + \varepsilon } )$ to avoid $\left\| {\bf{A}} \right\|_{2,1}^2{\rm{ = }}{\bf{0}}$. As a result, Eq.(\ref{eq:opti}) becomes  to Eq.(\ref{eq:S51})
    	\vspace{-5pt} 
	\begin{equation}\label{eq:S51}
		\resizebox{0.9\hsize}{!}{%
			$\begin{array}{l}
			{\bf{A'}} = \arg \mathop {\min }\limits_{\bf{A}} (tr({{\bf{A}}^T}{\bf{X}}({{\bf{M}}^{\rm{*}}}){{\bf{X}}^T}{\bf{A}}) + \alpha \left\| {\bf{A}} \right\|_F^2\\
			\,\,\,\,\,\,\, + \left\| {{{\bf{X}}^T}{\bf{A}} + {\bf{1}}{e^T} - {\bf{Y}}} \right\|_F^2 + \beta (\sum {_{j = 1}^d} \sqrt {\left\| {{{\bf{a}}^j}} \right\|_2^2 + \varepsilon } )
			\end{array}$}
	\end{equation}
	
	$\varepsilon $ is infinitely close to zero, which makes  Eq.(\ref{eq:S51}) closely equivalent to Eq.(\ref{eq:opti}).  Solving directly Eq.(\ref{eq:S51}) is non-trivial, we introduce a new variable ${\bf{G}} \in {R^{d*d}}$ which is a diagonal matrix with ${g_{jj}} = (\sum {_{i = 1}^d} \sqrt {\left\| {{{\bf{a}}^i}} \right\|_2^2 + \varepsilon } ) \div (\sqrt {\left\| {{{\bf{a}}^i}} \right\|_2^2 + \varepsilon } )$. $\textbf{G}$ and $\textbf{A}$ can be optimized iteratively. With  $\textbf{G}$ held fixed and  $\textbf{e}$  computed as in Eq.(\ref{eq:S42}), we can reformulate Eq.(\ref{eq:S51}) as Eq.(\ref{eq:S52})
    	\vspace{-5pt} 
	\begin{equation}\label{eq:S52}
		\resizebox{0.885\hsize}{!}{%
			$\begin{array}{l}
			{\bf{A'}} = \arg \mathop {\min }\limits_{\bf{A}} (tr({{\bf{A}}^T}{\bf{X}}({{\bf{M}}^{\rm{*}}}){{\bf{X}}^T}{\bf{A}}) + \alpha \left\| {\bf{A}} \right\|_F^2)\\
			\;\;\;\;\;\;\; + \left\| {{\bf{H}}{{\bf{X}}^T}{\bf{A}} - {\bf{HY}}} \right\|_F^2 + \beta tr({{\bf{A}}^T}{\bf{GA}})
			\end{array}$}
	\end{equation}
		\vspace{-5pt} 
	The closed form solution of Eq.(\ref{eq:S52}) is
    	\vspace{-1pt} 
	\begin{equation}\label{eq:S53}
		\resizebox{0.88\hsize}{!}{%
			${\bf{A'}} = {({\bf{X}}{{\bf{H}}^T}{\bf{H}}{{\bf{X}}^T} + \beta {\bf{G}} + \alpha {\bf{I}} + ({\bf{X}}({{\bf{M}}^{\rm{*}}}){{\bf{X}}^T}))^{ - 1}}{\bf{X}}{{\bf{H}}^T}{\bf{HY}}$}
	\end{equation}
		\vspace{-7pt} 
        
	\textit{\textbf{Step.5}} (Update of ${\bf{Y}}$)
	The label matrix \textbf{Y} contains two  parts: true labels ${{\bf{Y}}_S} = {\{ {y_1},...,{y_{{n_s}}}\} ^T} \in {{\bf{\mathbb{R}}}^{{n_s} \times c}}$, and pseudo labels  ${{\bf{Y}}_T} = {\{ {y_{{n_s} + 1}},...,{y_{{n_s} + {n_t}}}\} ^T} \in {{\bf{\mathbb{R}}}^{{n_t} \times c}}$. Our aim is to iteratively refine the latter ones. Given fixed $\textbf{A}$, $\textbf{e}$ and ${{\bf{M}}^*}$,  each ${y_{\rm{i}}} \in {{\bf{Y}}_T}$ can be updated by solving the following problem:
    	\vspace{-5pt} 
	\begin{equation}\label{eq:S61}
		\resizebox{0.7\hsize}{!}{%
			${y_i}^\prime  = \arg \mathop {\min }\limits_{{y_i} \ge 0,y_i^T{\bf{1}} = 1} \left\| {{{\bf{X}}^T}{\bf{A}} - {y_i}{\rm{ + }}{\bf{e}}} \right\|_F^2$}
	\end{equation}
	
	Using Lagrangian multipliers method, the final optimal solution of ${y_{\rm{i}}}$ is 
    	\vspace{-5pt} 
	\begin{equation}\label{eq:S62}
		\resizebox{0.4\hsize}{!}{%
			${y_i}^\prime  = ({{\bf{A}}^T}{{\bf{x}}_i} + e + \partial )$}
	\end{equation}
	where $\partial $ is coefficient of Lagrangian constraint $y_{\rm{i}}^T{\bf{1}} - 1 = 0$, which can be obtained by solving $y_{\rm{i}}^T{\bf{1}} = 1$.

	\textit{\textbf{Step.6}} (Update of ${{\bf{M}}^*}$)
		With labeled source data  ${{\bf{A}}^T}{{\bf{X}}_S}$ and pseudo labels generated on the target data ${{\bf{A}}^T}{{\bf{X}}_T}$ in \textbf{Step.5}, we can update  ${{\bf{M}}^*}$ as
        	\vspace{-5pt} 
		\begin{equation}\label{eq:S3}
		\resizebox{0.6\hsize}{!}{%
			${{\bf{M}}^*} = {{\bf{M}}_0} + \sum _{c = 1}^C({{\bf{M}}_c}) - {{\bf{M}}_{REP}}$}
		\end{equation}	
		where ${{{\bf{M}}_c}}$ and ${{{\bf{M}}_{REP}}}$ are defined in Eq.(\ref{eq:re}) and Eq.(\ref{eq:JDA}).

    	In summary, Algorithm 1 synthesizes the whole process for solving Eq.(\ref{eq:opti}). 
	
	\begin{algorithm}[!h]
		\caption{Discriminative  Label  Consistent Domain Adaptation (DLC-DA)}
		\KwIn{Data $\bf{X}$, Source domain label ${\bf{Y}}_{\cal S}$, subspace dimension $k$, iterations $T$, regularization parameters $\beta $ and $\alpha $}
		\textbf{{1}}: Initialize ${{\bf{M}}^*} = {{\bf{M}}_0}$ as defined in Eq.(\ref{eq:JDA}) ;\\
		\textbf{2}: Initialize $\textbf{A}$ by solving Eq.(\ref{eq:S2}); ($t: = 0$) \\
				\While{$\sim isempty(\bf{X},{{\bf{Y}}_{\cal S}})$ and $t<T$	}{
			\textbf{3}: Update ${{\bf{M}}^*}$ by solving Eq.(\ref{eq:S3}) \\
			\textbf{4}: Update ${\bf{e}}$ by solving Eq.(\ref{eq:S42})  \\
			\textbf{5}: Update ${\bf{A}}$; (${t_1}: = 0.$)\\
				\eIf{ ${t_1} < T$ }{
				(i) Initialize ${\bf{G}}$ as an identity matrix\;
				(ii) Update ${\bf{A}}$ by solving Eq.(\ref{eq:S53})\;
				(iii) Update ${\bf{G}}$ by calculating ${g_{jj}} = (\sum {_{i = 1}^d} \sqrt {\left\| {{{\bf{a}}^i}} \right\|_2^2 + \varepsilon } ) \div (\sqrt {\left\| {{{\bf{a}}^i}} \right\|_2^2 + \varepsilon } )$\;
				(iv) ${t_1} = {t_1} + 1$\;
			}{
			\textbf{break}\;
		}
		\textbf{6}: Update ${\bf{Y}}$ by solving Eq.(\ref{eq:S62})\\
		\textbf{7}: Update pseudo target labels $ {\bf{Y}}_{\cal T}^{(T)} = {{\bf{Y}}}\left[ {:,({n_s} + 1):({n_s} + {n_t})} \right] $;\\
		\textbf{8}:$t=(t+1)$;\\               
		
	}
	\KwOut{${\bf{A}}$, ${\bf{Z}} = {{\bf{A}}^T}{\bf{X}}$, $\textbf{Y}$}
\end{algorithm}

\subsection{Kernelization Analysis}
The proposed DLC-DA method can be extended to nonlinear problems in a Reproducing Kernel Hilbert Space via the kernel mapping $\phi :x \to \phi (x)$, or $\phi ({\bf{X}}):[\phi ({{\bf{x}}_1}),...,\phi ({{\bf{x}}_n})]$, and the kernel matrix ${\bf{K}} = \phi {({\bf{X}})^T}\phi ({\bf{X}}) \in {R^{n*n}}$. We utilize
	the Representer theorem to formulate Kernel DLC-DA as
	\begin{equation}\label{eq:kernel}
	\resizebox{1\hsize}{!}{%
		$\begin{array}{*{20}{l}}
		{\mathop {\min }\limits_{{{\bf{Y}}_{\bf{U}}} \ge 0,{{\bf{Y}}_{\bf{U}}}{\bf{1}} = {\bf{1}}} (tr({{\bf{A}}^T}{\bf{K}}{{\bf{M}}^*}{{\bf{K}}^T}{\bf{A}}) + \alpha \left\| {\bf{A}} \right\|_F^2 + \beta \left\| {\bf{A}} \right\|_{2,1}^2)}\\
		{\;\;\;\;\;\;\;\;\;\;\;\;\;\;\;\;\;\;\;\;\;\;\;\;\;\; + \left\| {{{\bf{K}}^T}{\bf{A}} - {\bf{1}}{{\bf{e}}^T} - {\bf{Y}}} \right\|_F^2}\\
		{{\rm{s}}t.{{\bf{M}}^*} = {{\bf{M}}_0} + \sum _{c = 1}^C({{\bf{M}}_c}) - {{\bf{M}}_{REP}},\;{\bf{Y}} \ge {\bf{0}},\;{\bf{Y1}} = {\bf{1}}}
		\end{array}$}
	\end{equation}

% \textcolor{red}{\subsection{Time Complexity Analysis}
% 	We analyze the computational complexity of Algorithm 1 using the big $O$ notation. We denote both the number of iteration $t,{t_1} \prec \min (m,n)$  and $ k \prec \min (m,n)$.
% 	The major computational burden of our algorithm lies in Steps 2, 4 and 6 presented in Section 3.3. In Step 2, the singular value decomposition is operated on $n*n$ matrix, the computational complexity is $O(k{n^2})$. In step 4, the matrix inversions computational complexity is $O({n^3})$. Step 6 constructs the MMMD matrix, which computational complexity is $O(4C{n^2})$. In Steps 2,5, $O(mn)$ for all other steps. Thus, the main computational complexity of Algorithm 1 is $O(t{t_1}{n^3} + 4tC{n^2} + tmn + k{n^2})$.}

\section{Experiments}
% In this section, we validate the effectiveness of our proposed  domain adaptation model, \textit{i.e.}, DLC-DA,  on 12 cross domain datasets for cross-domain image classification task. Results are then analyzed and compared to a series of state-of-the-art DA methods.

% In this section, we validate the effectiveness of our proposed  domain adaptation model, \textit{i.e.}, DLCDA,  on 16 cross domain datasets generated by permuting six datasets(Fig.2) for cross-domain image classification task. Results are compared to a series of state-of-the-art DA methods.

\subsection{Benchmarks}
In domain adaptation, Office+Caltech\cite{long2013transfer,DBLP:journals/tip/XuFWLZ16,DBLP:journals/tip/HouTYW16,DBLP:journals/ijcv/ShaoKF14} are standard benchmarks for the purpose of evaluation and comparison with state-of-the-art. In this paper, we follow the data preparation as most previous works\cite{DBLP:journals/tcyb/UzairM17,DBLP:journals/tip/XuFWLZ16,DBLP:journals/tip/HouTYW16,DBLP:conf/icml/GongGS13,DBLP:journals/pami/GhifaryBKZ17,DBLP:journals/tip/DingF17}. We construct 12 datasets for different image classification tasks.
	\vspace{-5pt} 
\begin{figure}[h!]
	\centering
	\includegraphics[width=1\linewidth]{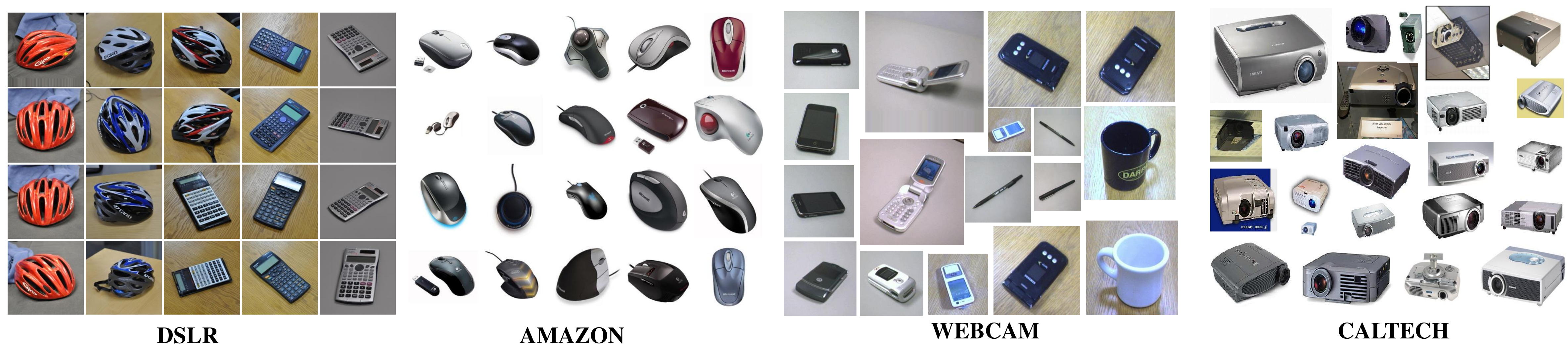}
	\caption { Sample images from the four datasets used in our experiments. Each dataset represents a different domain.} 
    \label{fig:data}
\end{figure} 			
	\vspace{-5pt}

\textbf{Office+Caltech} consists of 2533 images of ten categories (8 to 151 images per category per domain)\cite{DBLP:journals/pami/GhifaryBKZ17}, that forms four domains: (A) AMAZON, (D) DSLR, (W) WEBCAM, and (C) CALTECH. Fig.(\ref{fig:data}) illustrates some sample images from each domain.   We denote the dataset \textbf{Amazon},\textbf{Webcam},\textbf{DSLR},and \textbf{Caltech-256} by \textbf{A},\textbf{W},\textbf{D},and \textbf{C}, respectively.  $4\times 3=12$ domain adaptation tasks can then be constructed, namely  \emph{A} $\rightarrow$  \emph{W}  $\dots$ \emph{C} $\rightarrow$ \emph{D}, respectively.

Note that the arrow “$\rightarrow$” is the direction from “source” to “target”. For example, “Webcam $\rightarrow$ DSLR” means Webcam is the labeled source domain while DSLR  the unlabeled target.

\subsection{State of the art DA Methods}
The proposed DLCDA method is compared with \textbf{twenty-three} methods of the literature, including deep-learning based approaches for unsupervised domain adaption, given the fact that we also made use of deep features in our experiments. They are: 
(1)1-Nearest Neighbor Classifier (\textbf{NN}); 
(2) Principal Component Analysis (\textbf{PCA}); 
(3) \textbf{GFK}\cite{gong2012geodesic}; 
(4)\textbf{TCA}\cite{pan2011domain}; 
(5)\textbf{TSL}\cite{4967588}; 
(6)\textbf{JDA}\cite{long2013transfer}; 
(7)\textbf{ELM}\cite{DBLP:journals/tcyb/UzairM17}; 
(8)\textbf{AELM}\cite{DBLP:journals/tcyb/UzairM17}; 
(9)\textbf{SA}\cite{DBLP:conf/iccv/FernandoHST13}; 
(10)\textbf{mSDA}\cite{DBLP:journals/corr/abs-1206-4683}; 
(11)\textbf{TJM}\cite{DBLP:conf/cvpr/LongWDSY14}; 
(12)\textbf{RTML}\cite{DBLP:journals/tip/DingF17}; 
(13)\textbf{SCA}\cite{DBLP:journals/pami/GhifaryBKZ17}; 
(14)\textbf{CDML}\cite{DBLP:conf/aaai/WangWZX14}; 
(15)\textbf{DDC}\cite{DBLP:journals/corr/TzengHZSD14}; 
(16)\textbf{LTSL}\cite{DBLP:journals/ijcv/ShaoKF14}; 
(17)\textbf{LRSR}\cite{DBLP:journals/tip/XuFWLZ16}; 
(18)\textbf{KPCA}\cite{DBLP:journals/neco/ScholkopfSM98}; 
(19)\textbf{JGSA} \cite{Zhang_2017_CVPR}; 
(20)\textbf{DAN}\cite{long2015learning};
(21)\textbf{AlexNet}\cite{krizhevsky2012imagenet}
(22)\textbf{PUnDA}\cite{Gholami_2017_ICCV}
(23)\textbf{TAISL}\cite{Lu_2017_ICCV}.

% In addition, for the purpose of fair comparison, we also follow the experimental settings of \textbf{JDA}, \textbf{LTSL}, \textbf{DAN} and \textbf{LRSR}, and apply DeCAF6 as the features for some methods to be evaluated. Whenever possible, we directly reported the performance scores from the publications, \textit{i.e.},  \cite{long2013transfer,DBLP:journals/tcyb/UzairM17,DBLP:journals/tip/DingF17,DBLP:journals/pami/GhifaryBKZ17,DBLP:journals/tip/XuFWLZ16,Zhang_2017_CVPR,Gholami_2017_ICCV} ,   of the \textbf{twenty-four} methods listed above. They are assumed to be their \emph{best} performance.

\subsection{Experimental Setup}
 We used two types of  features extracted from these datasets that are publicly available, namely \textbf{SURF} and \textbf{DeCAF6} features. The \textbf{SURF}\cite{gong2012geodesic} features are extracted and quantized into an 800-bin histogram with the  codebook computed with Kmeans on a subset of images from Amazon. Then the histograms are standardized by z-score. \textbf{Deep Convolutional Activation Features (DeCAF6)}\cite{DBLP:conf/icml/DonahueJVHZTD14} were constructed  as in previous research\cite{DBLP:journals/pami/GhifaryBKZ17,Zhang_2017_CVPR,Lu_2017_ICCV} which uses the VLFeat MatConvNet\cite{DBLP:journals/corr/VedaldiL14} library with a number of pretrained CNN models. With the proposed  Caffe\cite{DBLP:conf/mm/JiaSDKLGGD14} implementation of AlexNet\cite{DBLP:journals/cacm/KrizhevskySH17}  trained on the ImageNet dataset, we used the outputs from the 6th layer as the features, leading to 4096 dimensional \textbf{DeCAF6} features. 
 
The proposed \textbf{DLC-DA}  involves three hyper-parameters: subspace dimension $k$  and two regularization parameters, \textit{i.e.},  $\alpha $ and $\beta $ as defined in Eq.(\ref{eq:opti}) and Eq.(\ref{eq:S2}). In this experiment, we set $k = 100,\;\alpha  = 1,\;\beta  = 1.1$. In section 4.4.2, we also analyzes the sensitivity of the proposed DLC-DA with respect to these three parameters. 

% The {\emph{accuracy}}  on the test dataset, as defined in Eq.(\ref{eq:accuracy}),  is the evaluation measurement. It is widely used in literature, \textit{e.g.},\cite{pan2008transfer,long2015learning,DBLP:journals/corr/LuoWHC17,long2013transfer,DBLP:journals/tip/XuFWLZ16}, \textit{etc}.

% \begin{equation}\label{eq:accuracy}
% 	\begin{array}{c}
% 		Accuracy = \frac{{\left| {x:x \in {D_T} \wedge \hat y(x) = y(x)} \right|}}{{\left| {x:x \in {D_T}} \right|}}
% 	\end{array}
% \end{equation}
% where ${\cal{D_T}}$ is the target domain, ${\hat{y}(x)}$ is the predicted label and ${y(x)}$ is the ground truth label for a test data  $x$.

\subsection{ Results and Discussion}

\subsubsection{Experiments on the Office+Caltech-256 Data Sets}

% Comparison with the state-of-the-art DA methods is made using two settings. The first one makes use of the SURF features (Fig.\ref{fig:accSO}) whereas the second one utilizes the deep  DeCAF6 features (Fig.\ref{fig:accDO}), and thus enables the proposed DLCDA method to be compared with deep learning-based DA methods.  
We  follow the experimental settings of \textbf{JDA}, \textbf{LTSL}, \textbf{DAN} and \textbf{LRSR}, and apply DeCAF6 as the features for some methods to be evaluated. However, whenever possible in these figures, we directly report the performance scores from the publications, \textit{i.e.},  \cite{long2013transfer,DBLP:journals/tcyb/UzairM17,DBLP:journals/tip/DingF17,DBLP:journals/pami/GhifaryBKZ17,DBLP:journals/tip/XuFWLZ16,Zhang_2017_CVPR,Gholami_2017_ICCV} ,   of the \textbf{twenty-three} methods listed above. They are assumed to be their \emph{best} performance.

\begin{figure}[h!]
	\centering
	\includegraphics[width=1\linewidth]{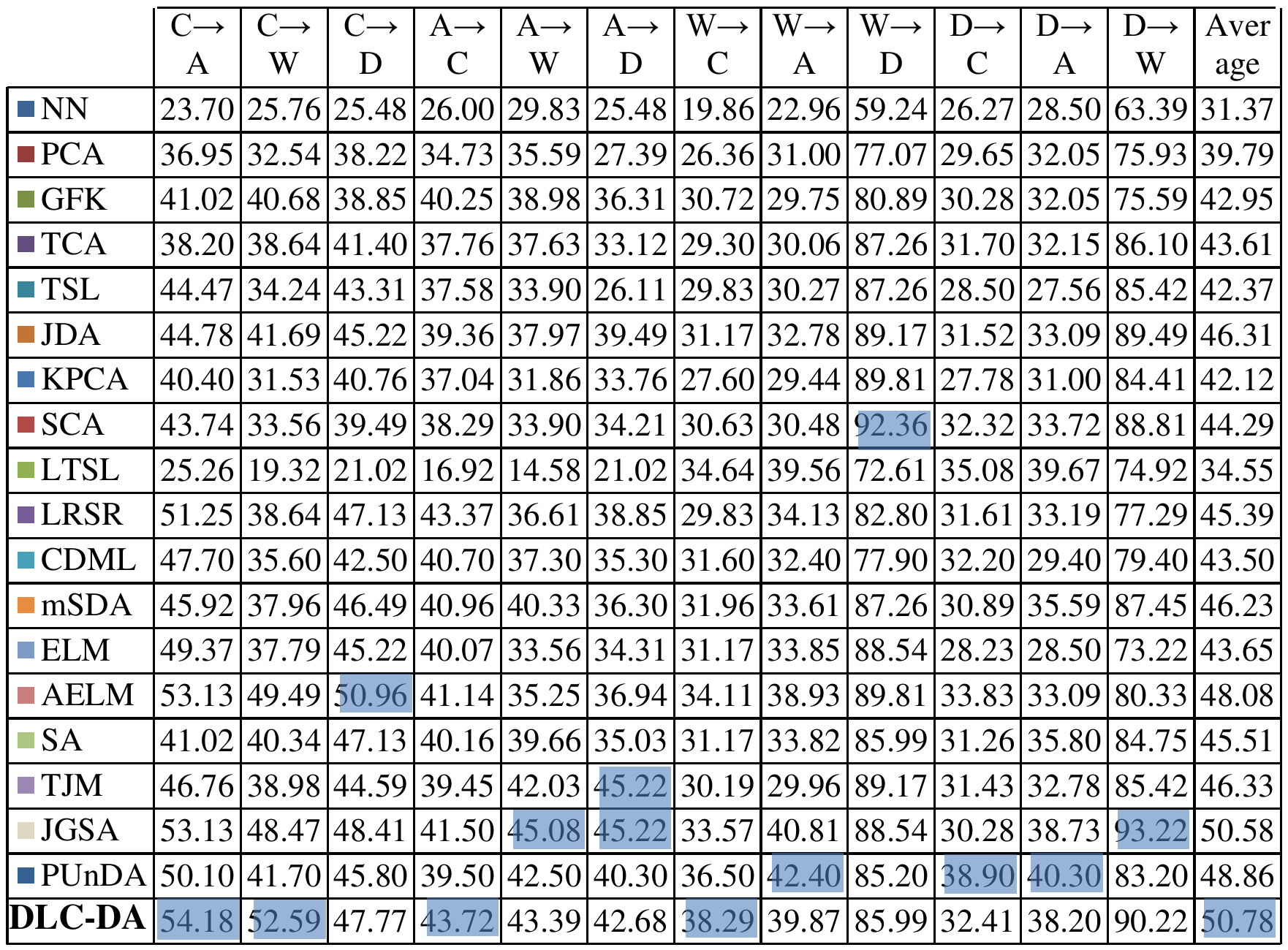}
	\caption{Accuracy(${\rm{\% }}$) on the Office+Caltech Images with SURF-BoW Features} 
    	\label{fig:accSO}
\end{figure}

\begin{figure}[h!]
	\centering
	\includegraphics[width=1\linewidth]{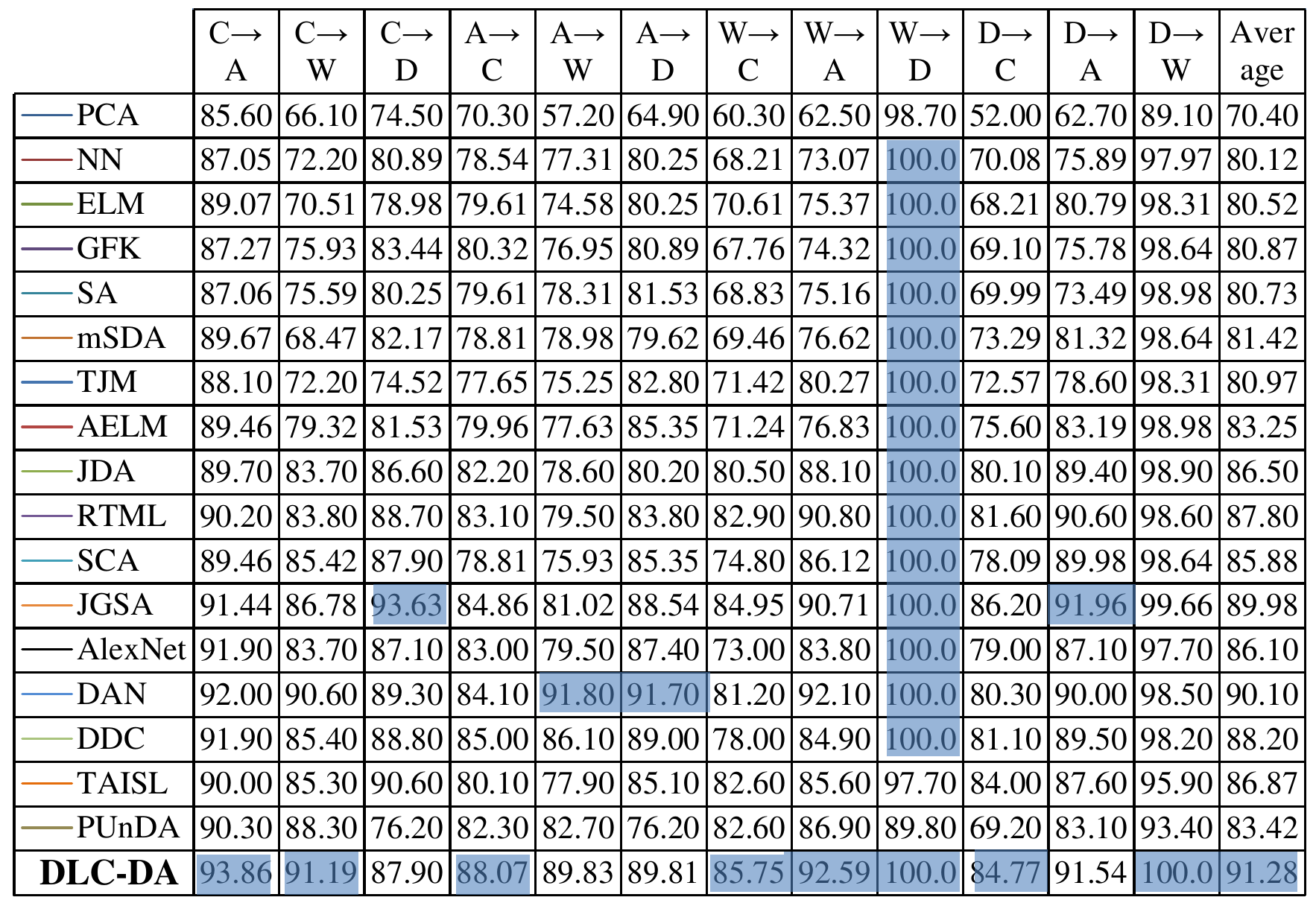}
	\caption {Accuracy(${\rm{\% }}$) on the Office+Caltech Images with the DeCAF6 Features
    	\label{fig:accDO}
	} 
\end{figure}

As can be seen from Fig.\ref{fig:accSO} and Fig.\ref{fig:accDO}, the experimental results verify the effectiveness of the proposed DLC-DA method which consistently outperforms state-of-the-art DA methods, whether with the traditional shallow SURF features or the deep DeCAF6 features. However, given such a performance,  one natural question that one can raise is how the proposed method is sensitive to the choice of the hyper-parameters. This issue is analyzed in the next subsection.   

% In Tab.1 and Tab.2, our methods rank first performance on average accuracy which shows the effectiveness of the proposed method in domain adaptation. Moreover, DRDA and CDDA are only two methods which achieves full accuracy ($100\% $) both in cases, e.g.,  \emph{W} $\rightarrow$ \emph{D} $ and $ \emph{D} $\rightarrow$ \emph{W}. 	Compare with Tab.1 and Tab.2, the transfer learning perfromance of DRDA would be highly improved through increasing the quality of features. It is important to emphasize that the proposed DRDA method is not a feature extraction technique but a DA technique. The perfromance of DRDA could be imporved if we extract more efficient features.

\subsubsection{Parameter Sensitivity Analysis}
\begin{figure}[h!]
	\centering
	\includegraphics[width=1\linewidth]{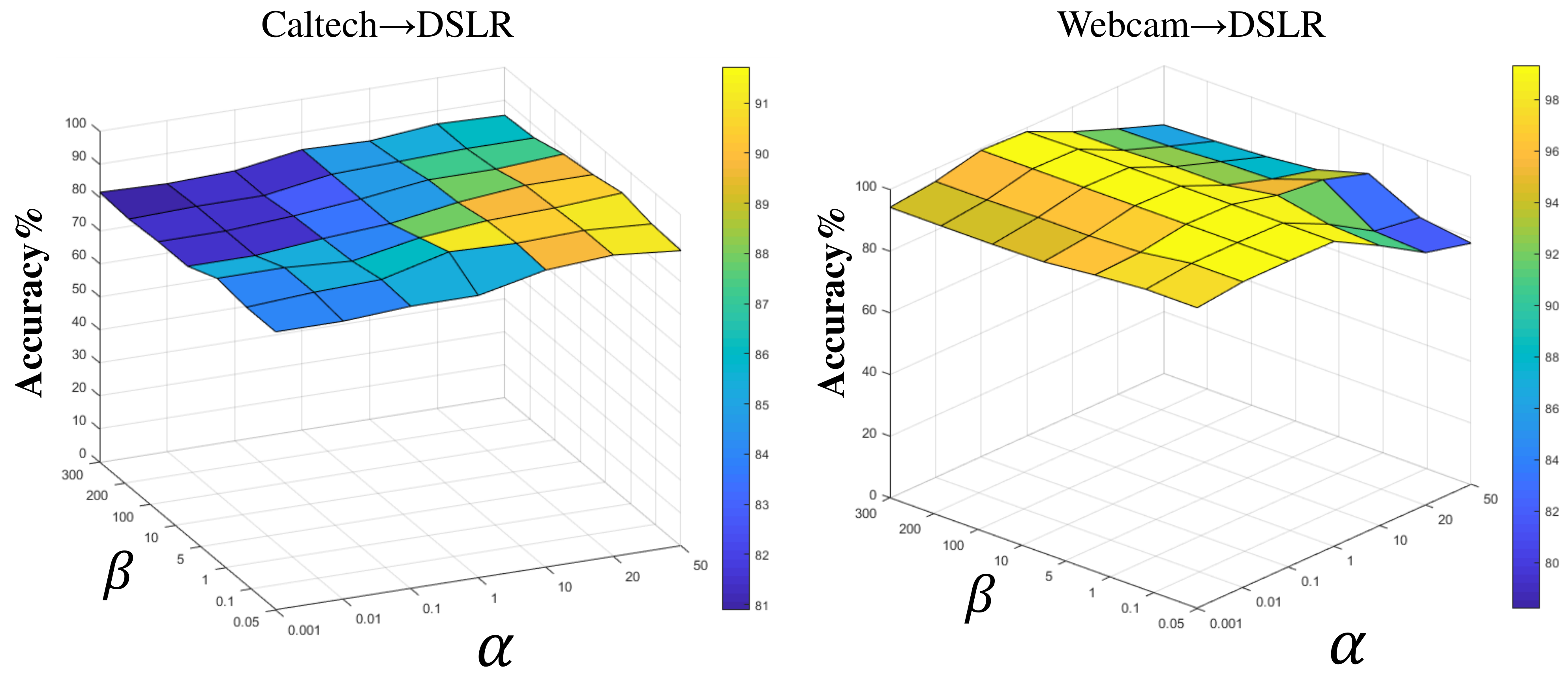}
	%	\vspace{-7pt}
	\caption { The classification accuracies of the proposed DLC-DA method vs. the parameters $\alpha $ and $\beta $ on the selected three cross domains data sets, with $k$ held fixed at $100$.} 
    	\label{fig:para}
\end{figure}

 $\alpha $ and $\beta $ as defined in Eq.(\ref{eq:opti}) are the major hyper-parameters of the proposed DLC-DA method. While  $\alpha $ aims to regularize the projection matrix $A$ to avoid over-fitting the chosen shared feature subspace with respect to both source and target data, $\beta$ as expressed in Eq.(\ref{eq:RLR}) controls the dimensionality of class dependent data manifold in the searched shared feature subspace, or in other word the sparsity level of the linear combination of the projected features to regress the class label. We study the sensitivity of the proposed DLC-DA method with a wide range of parameter values, \textit{i.e.}, $\alpha  = (0.001,0.01,0.1,1,10,20,50)$ and $\beta  = (0.05,0.1,1,5,10,100,200)$. We only report the results on \emph{C} $\rightarrow$ \emph{D} $ and $ \emph{W} $\rightarrow$ \emph{D} datasets due to the space limitation. With $k$ held fixed at $100$,  Fig.\ref{fig:para} illustrates these results. As can be seen from Fig.\ref{fig:para}, the proposed DLC-DA displays its stability as the resultant classification  accuracies remain roughly the same despite  a wide range of $\alpha $ and $\beta $ values. 

%  Thus, it is a easy job to select a suitable  parameter combination for our method.
 
%  We evaluate different combinations of these values selected from a reasonable discrete set $\alpha  = (0.001,0.01,0.1,1,10,20,50)$ and $\beta  = (0.05,0.1,1,5,10,100,200)$ on three cross domains data sets. 
	
\begin{figure}[h!]
	\centering
		\includegraphics[width=1\linewidth]{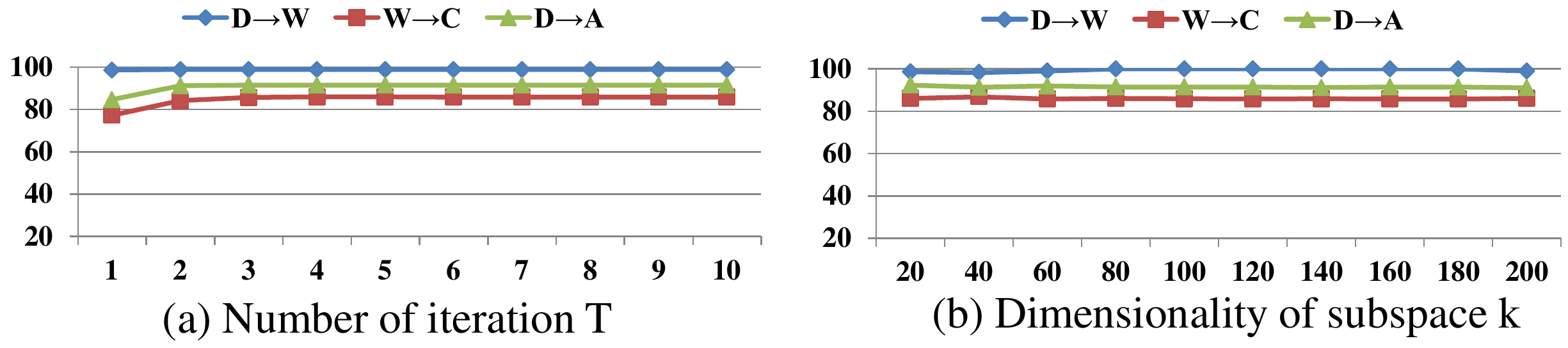}
	%	\vspace{-7pt}
	\caption{Sensitivity analysis of the proposed DLC-DA method \textit{w.r.t.} (a): the number of iterations $T$, and (b): the subspace dimensionality $k$, using the Decaf6 features over the three  datasets, \textit{i.e.}, DSLR (D), Webcam (W), and Caltech (C).} 
    \label{fig:iter}
\end{figure}

In Fig.\ref{fig:iter}, We further perform a convergence analysis of the proposed DLC-DA method, namely the performance convergence \textit{w.r.t.} the number of iterations $T$, as well as the impact of the chosen dimensionality, \textit{i.e.}, $k$,  of the searched shared feature subspace. We report the results on \emph{D} $\rightarrow$ \emph{W},  \emph{W} $\rightarrow$ \emph{C} $ and $  \emph{D} $\rightarrow$ \emph{A}. In Fig.\ref{fig:iter}.a, we vary the number of iterations $T = (1,2,3,4,5,6,7,8,9,10)$, whereas the subspace dimensionality $k$ varies with $k = (20,40,60,80,100,120,140,160,180,200)$  in Fig.\ref{fig:iter}.b.   
% Fig.5a and Fig.5b illustrates the relationship between iteration $T = (1,2,3,4,5,6,7,8,9,10)$ ; dimensionality of subspace $k = (20,40,60,80,100,120,140,160,180,200)$ and the accuracy respectively. 
As can be seen from Fig.\ref{fig:iter}.a, the proposed DLC-DA achieves its optimal performance only after 2 iterations. Furthermore, Fig.\ref{fig:iter}. shows that DLC-DA remains stable \textit{w.r.t.} a wide rang of $k \in [20,200]$.

\vspace{-5pt}
\subsubsection{Analysis and Verification}
\begin{figure}[h!]
	\centering
	\includegraphics[width=1\linewidth]{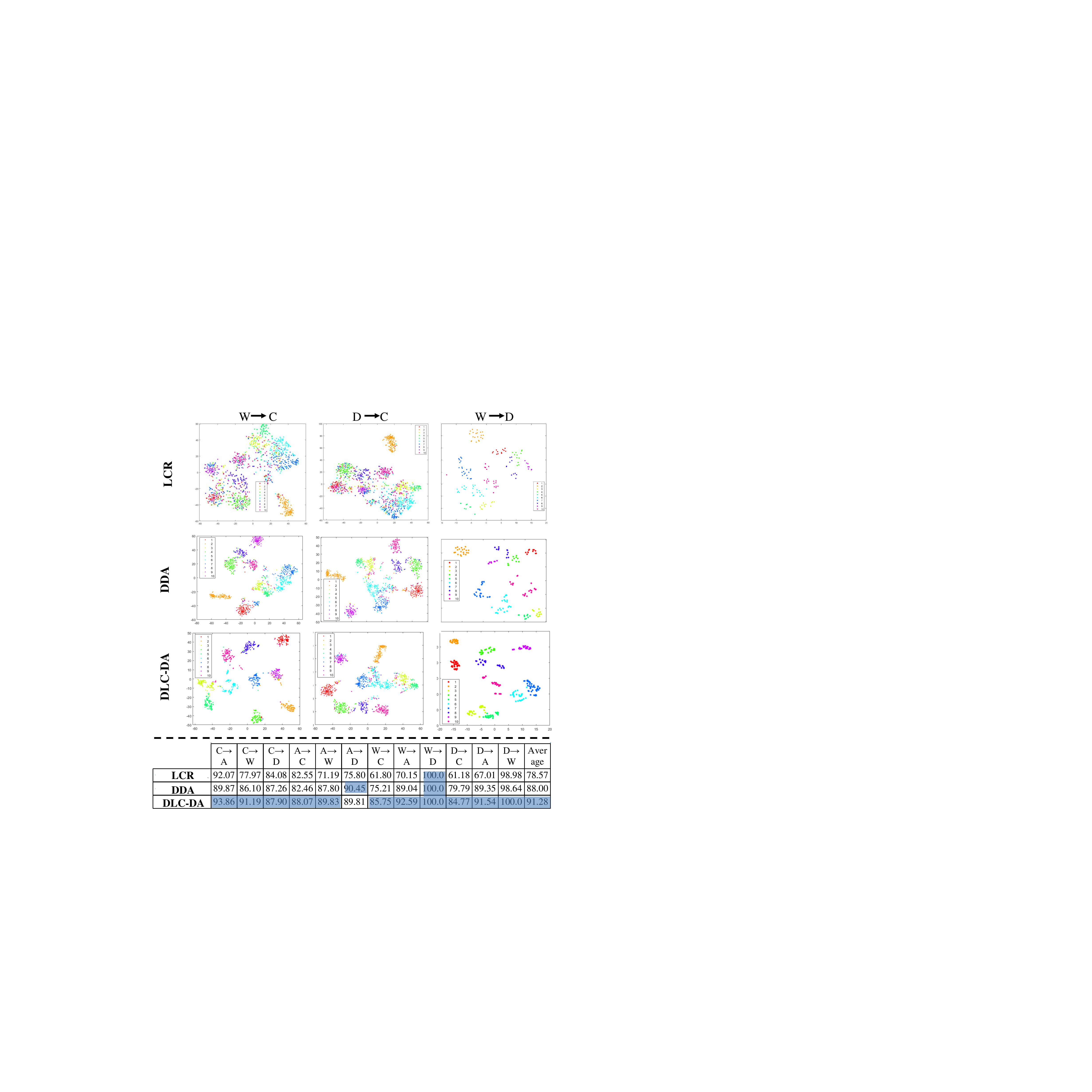}
	\caption {Accuracy(${\rm{\% }}$) on the Office+Caltech Images with DeCAF6 Features}
    	\label{fig:VI}
\end{figure}

\vspace{-2pt}
The core model of the proposed DLC-DA method is Eq.(\ref{eq:opti}) which adds up two optimization terms, namely discriminative data distribution alignment (DDA) term as defined in Eq.(\ref{eq:DLCDA-R}), and label consistent regression (LCR) term as defined in Eq.(\ref{eq:RLR}). The DDA and LCR term aim to decrease discriminatively data distribution mismatch and ensure label consistency as defined by the second and third term of the error bound in Eq.(\ref{eq:bound}). One interesting question is how each of these terms contributes to the proposed final model as defined by Eq.(\ref{eq:opti}). For this purpose, we derive from Eq.(\ref{eq:opti}) two additional partial DLC-DA methods, namely DLC-DA(DDA) making only use of discriminative distribution alignment as defined in Eq.(\ref{eq:DLCDA-R}) and DLC-DA(LCR) restricted to label consistent regression as defined in Eq.(\ref{eq:RLR}). They are benchmarked using the \textbf{Office+Caltech} datasets with the \textbf{DeCAF6} features in comparison with the proposed DLC-DA method. Fig.\ref{fig:VI} plots these experimental results.

% In this section, three experiments will be designed to verify the contributions of designed constraints. To save the limited space, experiments are only proposed on the \textbf{Office+Caltech} data sets with \textbf{DeCAF6} features.

% The goal of the first two experiments in Fig.3 is to evaluate whether
% the joint optimization of distribution approximation and linear regression really boost the
% classification performance. To achieve this goal, the \textbf{DLC-DA-R}
% is designed for testing \textbf{DLC-DA} without the linear regression constraints and \textbf{DLC-DA-D} tries to test effectiveness of distribution approximation constraints. Specifically, the objective function of \textbf{DLCDA-R} is:
\begin{equation}\label{eq:DLCDA-R}
	\begin{array}{l}
		\min (tr({{\bf{A}}^T}{\bf{X}}{{\bf{M}}^*}{{\bf{X}}^T}{\bf{A}}) + \alpha \left\| {\bf{A}} \right\|_F^2)\\
		{\rm{s}}t.{{\bf{A}}^T}{\bf{XH}}{{\bf{X}}^T}{\bf{A}} = {\bf{I}}
	\end{array}
\end{equation}

% The objective function of \textbf{DLC-DA-D} is
% \begin{equation}\label{eq:DRDA-D}
% 	\begin{array}{*{20}{l}}
% 		{\min \beta \left\| {\bf{A}} \right\|_{2,1}^2 + \left\| {{{\bf{X}}^T}{\bf{A}} - {\bf{1}}{{\bf{e}}^T} - {\bf{Y}}} \right\|_F^2}\\
% 		{{\rm{s}}t.\;{\bf{Y}} \ge {\bf{0}},\;{\bf{Y1}} = {\bf{1}}}
% 	\end{array}
% \end{equation}

% The objective function of \textbf{DLC-DA} is Eq.(\ref{eq:opti}). 

As can be seen from this figure, the proposed \textbf{DLC-DA} outperforms \textbf{DLC-DA(LCR)} by ${\bf 12}\uparrow$ points and  \textbf{DLC-DA(DDA)} by ${\bf 3}\uparrow$ points. These results thus suggest the complementarity of the DDA and LCR terms and the added-value of their joint optimization. To gain intuition, Fig.\ref{fig:VI} further visualizes class explicit data distributions in the resultant shared feature subspace using the three variants of DLC-DA over the three cross-domain datasets, namely \emph{W} $\rightarrow$ \emph{C} , \emph{D} $\rightarrow$ \emph{C}  and \emph{W} $\rightarrow$ \emph{D}. Different colors represent different classes. As can be seen in the figure,  DLC-DA(DDA) shows its effectiveness in compacting intra-class instances. When the DDA term is further combined with the LCR term, the proposed DLC-DA better separates data from different classes in increasing inter-class distances. 

\begin{figure}[h!]
	\centering
	\includegraphics[width=1\linewidth]{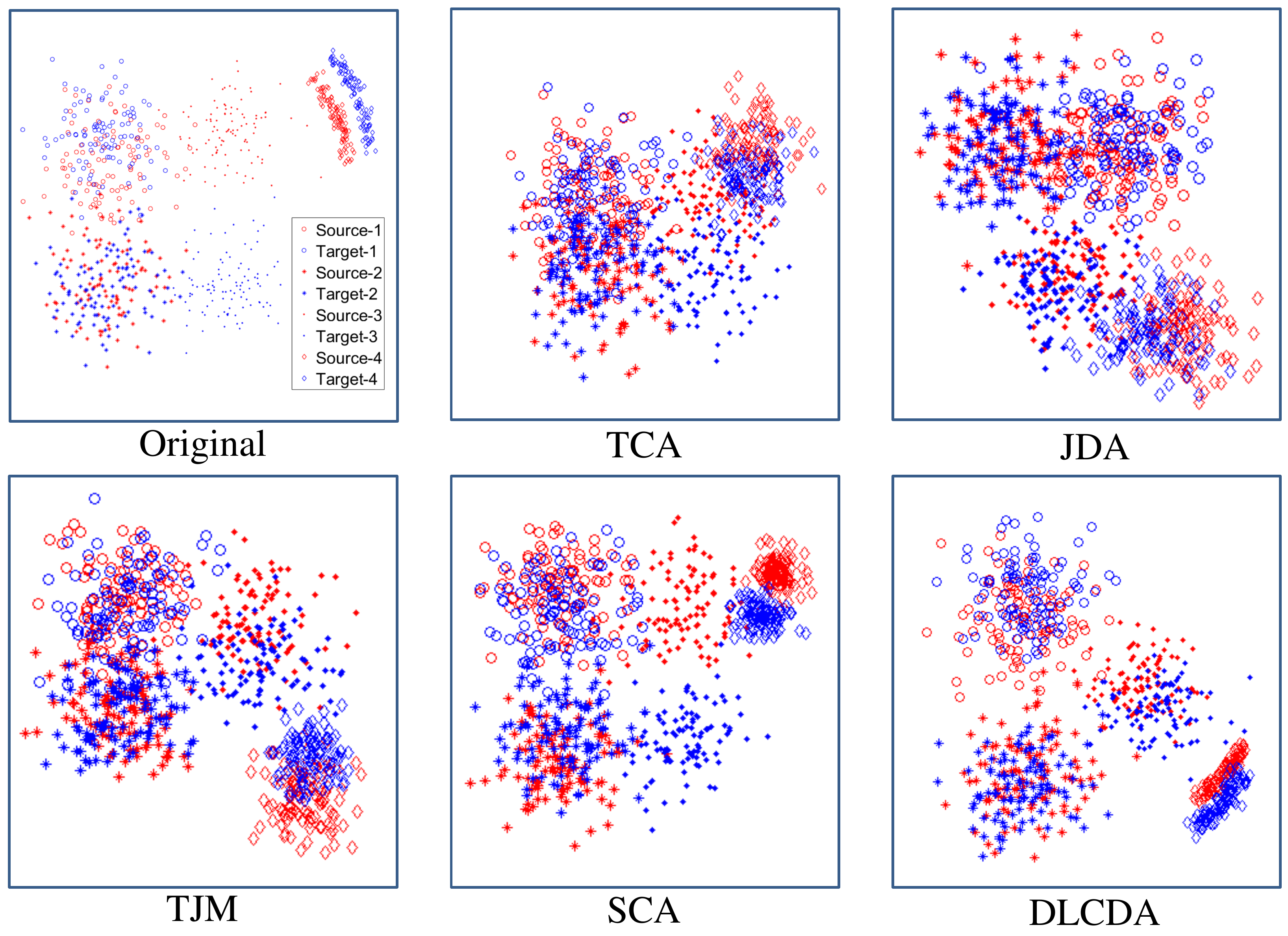}
	\caption {Comparisons of baseline domain adaptation methods and the proposed DLCDA method on the synthetic data	}
    	\label{fig:compare}
\end{figure} 	

To further gain insight of the proposed DLC-DA \textit{w.r.t.} its domain adaptation skills, we also evaluate DLC-DA using a synthetic dataset in comparison with several state of the art DA methods. Fig.\ref{fig:compare} visualizes the original data distributions with 4 classes and the resultant shared feature subspaces as computed by TCA, JDA, TJM, SCA and DLC-DA, respectively. In this experiment, we focus our attention on the ability of the DA methods to align discriminatively data distributions between source and target. As such, the original synthetic data depicts slight distribution discrepancies between source and target for the first two class data, wide distribution mismatch for the third and fourth class data. Fourth class data further depict a moon like geometric structure. As can be seen in Fig.\ref{fig:compare}, baseline methods have difficulties to align data distributions with wide discrepancies, \textit{i.e.}, third or fourth class data. In contrast, thanks to the joint use of the DDA and LCR terms, the proposed DLC-DA not only align data distributions compactly but also separate class data very distinctively.

\section{Conclusion}
We have proposed in this paper a novel unsupervised DA method, namely Discriminative Label Consistent Domain Adaptation (DLC-DA), which, in contrast to state of the art DA methods only focused on data alignment, simultaneously optimizes three terms of the upper error bound of a learned classifier on the target domain. Furthermore, data outliers are also explicitly accounted for in our model to avoid negative transfer. Comprehensive experiments using the standard Office+CaltechA benchmark in DA show the effectiveness of the proposed method which consistently outperform state of the art DA methods. 

% in searching  iteratively a shared feature subspace where data distributions between the source and target domain are discriminatively aligned whereas at the same time data labels are sparsely regressed from the features achieved in the latent shared subspace.      
{\small
	\bibliographystyle{ieee}
	\bibliography{egbib}
}

\end{document}